\documentclass[letterpaper, 10 pt, journal, twoside]{ieeetran}

\usepackage{cite}

\usepackage[
    colorlinks=true,
    linkcolor=blue,
    citecolor=blue,
    urlcolor=black
]{hyperref}
\usepackage{amsmath,amssymb,amsfonts}
\usepackage{graphicx}
\usepackage{textcomp}

\usepackage[table]{xcolor}
\usepackage{orcidlink}
\usepackage{algorithm}
\usepackage{algpseudocode}
\usepackage{float}
\usepackage{booktabs}
\usepackage{multirow}
\usepackage{wrapfig}
\usepackage{tabularx}
\usepackage{array}
\usepackage{ragged2e}
\usepackage{threeparttable}
\usepackage{adjustbox}
\usepackage{siunitx}
\usepackage{makecell}
\usepackage{dblfloatfix}
\usepackage{rotating}
\usepackage{titlesec}

\usepackage{pifont}
\usepackage{xcolor}

\newcommand{\cmark}{\textcolor{green!60!black}{\ding{51}}}
\newcommand{\xmark}{\textcolor{red}{\ding{55}}}

\definecolor{checkgreen}{RGB}{0,130,70}
\definecolor{crossred}{RGB}{200,40,40}

\title{\LARGE \bf
HexaGripper: A Single-Actuator, Winch-Deployed Gripper for Autonomous Aerial Parcel Collection
}

\author{
Nimantha Adikaram$^{1,\dagger}$\,\orcidlink{0009-0005-2064-1555},
Mahen Abeyratne$^{1,\dagger}$\,\orcidlink{0009-0002-3255-5599},
Lakmina Chandrajith$^{1}$\,\orcidlink{0009-0004-8564-1057},\\
A.~H.~T.~E. De Silva$^{1}$\,\orcidlink{0000-0003-0518-1843},
Isira Naotunna$^{2}$\,\orcidlink{0000-0003-0736-4483},
and Asanka Perera$^{3}$\,\orcidlink{0000-0003-4021-3943}%
\thanks{This work has been submitted to the IEEE for possible publication.
Copyright may be transferred without notice, after which this version
may no longer be accessible.}%
\thanks{$^{1}$Department of Mechanical Engineering, University of Moratuwa,
Moratuwa 10400, Sri Lanka.
{\tt\small \{adikaramwkdne.21, abeyratnaudmp.21, chandrajithbpl.21, ahteranga\}@uom.lk}}%
\thanks{$^{2}$School of Electrical and Mechanical Engineering,
The University of Adelaide, Adelaide, SA, Australia.
{\tt\small isira.naotunna@adelaide.edu.au}}%
\thanks{$^{3}$School of Science, Engineering and Digital Technologies,
University of Southern Queensland, Australia.
{\tt\small Asanka.Perera@unisq.edu.au}}%
\thanks{$^{\dagger}$These authors contributed equally to this work.}%
}

\begin{document}

\maketitle

\begin{abstract}
Autonomous aerial parcel collection remains constrained by package-transfer operations that require manual loading, landing, or dedicated ground infrastructure. This letter presents HexaGripper, a single-actuator, winch-deployed six-plate gripper for autonomous collection of cuboid parcels. The proposed mechanism provides synchronized grasping with a wide capture region and tolerance to positional misalignment during pickup. The system integrates vision-based alignment, range sensing, winch deployment, and state-based control to enable autonomous collection while maintaining UAV separation from the pickup surface. Experimental evaluation covered static grasping, capture-workspace assessment, and end-to-end outdoor aerial collection. Static experiments achieved 100\% pickup success (15/15 trials) across three parcel geometries, while workspace experiments achieved successful pickup at all 28 tested positions up to a 150~mm radial offset. End-to-end outdoor aerial collection achieved an 80\% success rate (4/5 trials). These results demonstrate the feasibility of mechanically synchronized, winch-deployed grasping for landing-free autonomous aerial parcel collection.
\end{abstract}

\section{Introduction}

Parcel-delivery networks increasingly require rapid transportation in dense urban areas, remote communities, and locations with limited road access. Unmanned Aerial Vehicles (UAVs) can reduce dependence on ground infrastructure; however, parcel pickup and transfer often still require human intervention or dedicated infrastructure~\cite{Zieher2024}. Existing delivery
approaches primarily rely on packages attached before takeoff~\cite{Porghoveh2026}, landing-based
transfer~\cite{Smith2023,Domislovic2025, Cheung2024}, tethered lowering~\cite{Kang2025}, or dedicated docking infrastructure~\cite{Porghoveh2026, Domislovic2025, Kang2025}. Collecting or dispatching a package
requires a suitable surface and brings the aircraft close to obstacles and it
makes the UAV more vulnerable to collisions, whereas hover and tethered delivery maintains separation from
the parcel but generally does not address autonomous collection and dispatching.
A self-loading and unloading UAV must therefore establish and verify a secure mechanical
connection with a parcel despite positioning error.

\begin{figure}[t]
    \centering
    \includegraphics[width=\columnwidth]{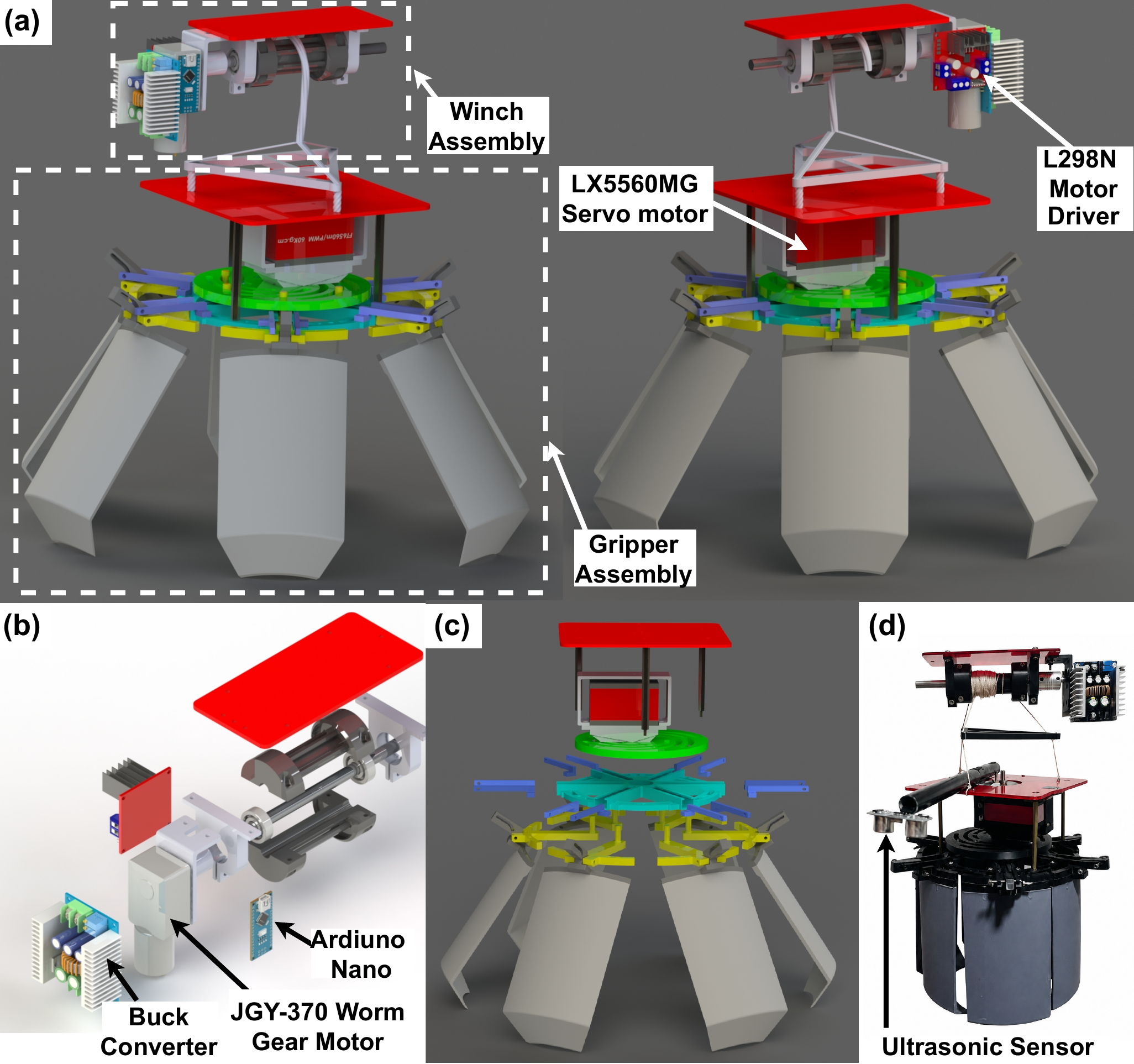}
    \caption{HexaGripper system design and prototype:
    (a) integrated gripper-winch assembly and major components,
    (b) exploded view of winch assembly, (c) exploded view of the gripper assembly, and
    (d) fabricated physical prototype.}
    \label{fig:hexagripper-assembly}
\end{figure}

Aerial manipulation has been investigated using several mechanism classes.
Multi-finger robotic hands~\cite{Zhao2025,Nishimura2026,Setty2020} provide broad object compatibility but introduce additional actuators, sensing, and contact-control requirements. Passive bio-inspired grippers~\cite{Firouzeh2024} can reduce active control and achieve rapid mechanically triggered closure, while foldable aerial manipulators~\cite{Wang2025} provide compact stowage and extended reach at the cost of additional articulated degrees of freedom and endpoint positioning requirements. Compliant and tentacle-based grippers~\cite{Yang2024, Cheung2024} can accommodate irregular object geometries, although pneumatic or multi-arm implementations introduce additional hardware and packaging
requirements. In~\cite{Appius2022}, rapid aerial pickup was demonstrated using a specialized capture mechanism, and linkage-based aerial grippers~\cite{Smith2023} have demonstrated the use of mechanical synchronization and sensing to simplify aerial grasping. For tethered manipulation, in \cite{Kang2025}, the locomotion of the end-effector is used to correct the positioning error after the deployment. Overall, these approaches involve tradeoffs among object adaptability, actuation complexity, positioning tolerance, and payload capacity. However, autonomous parcel collection and dispatch introduces several additional requirements. The gripper must tolerate UAV positioning errors, vibration, securely enclose parcels, and maintain low actuation complexity. Tethered deployment enables parcel collection and dispatch while maintaining separation
between the UAV and the pickup surface. These requirements motivated the authors to design mechanically mechanically synchronized gripping elements with a wide capture region. Given the UAV's 6.5~kg payload capacity, a maximum mass of 1.5~kg was allocated to the tethered-winch subsystem. The mechanism was also required to fit within the landing-gear envelope when stowed and retain the parcel without continuous winch actuation.

\begin{figure}[t]
    \centering
    \includegraphics[width=\columnwidth]{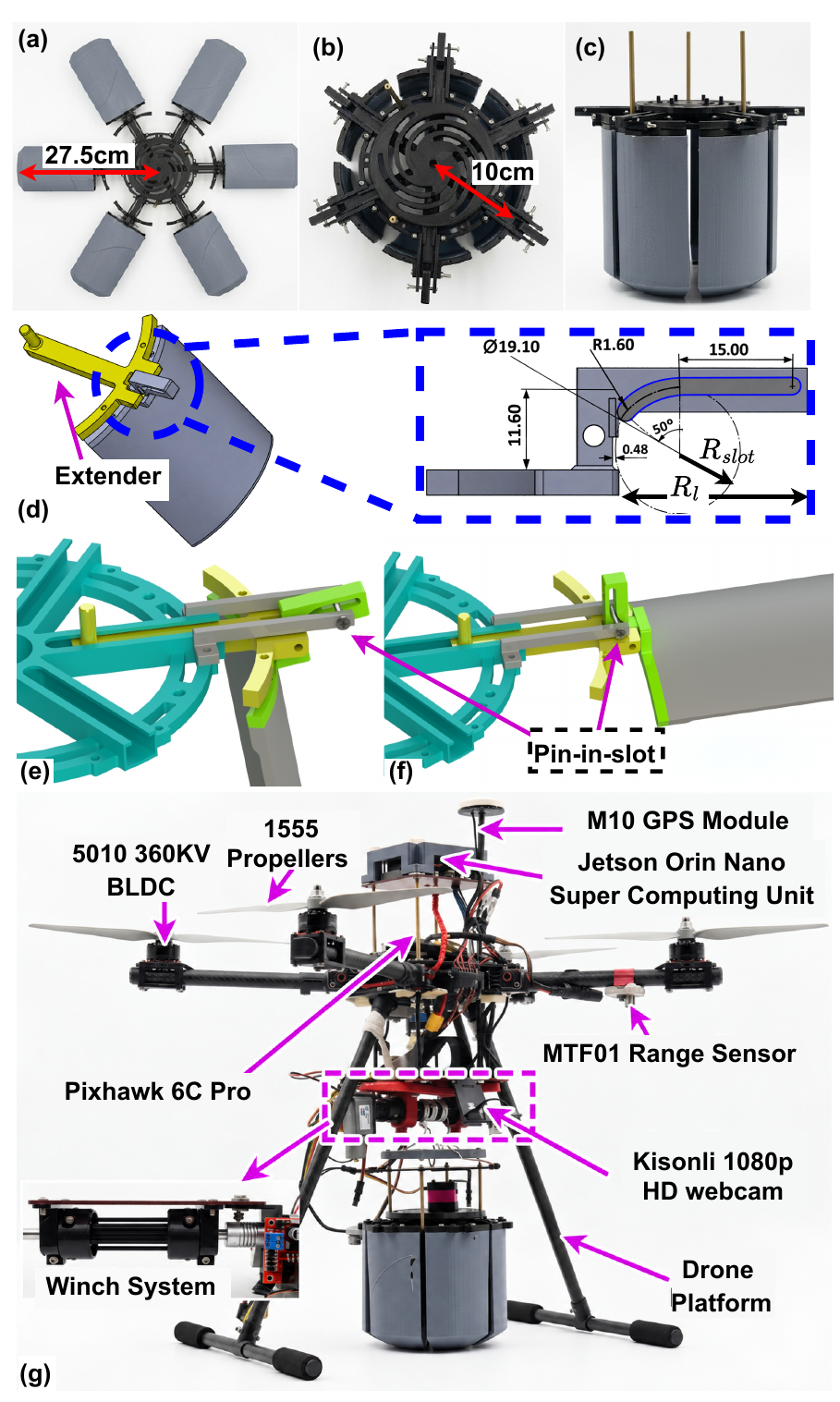}
    \caption{HexaGripper prototype and UAV integration:
    (a) open six-plate configuration,
    (b) central cam and radial-extender mechanism,
    (c) assembled HexaGripper,
    (d) plate, connector, radial extender, and follower assembly showing
    the pin-in-slot mechanism,
    (e) initial plate configuration,
    (f) fully extended configuration after $90^\circ$ plate rotation, and
    (g) integration with the UAV platform.}
    \label{fig:prototype}
\end{figure}

This letter presents HexaGripper (Fig.~\ref{fig:hexagripper-assembly}), a
winch-deployed mechanism for autonomous collection of approximately cuboid
parcels. A single rotary cam synchronizes six gripping plates, while a
pin-in-slot mechanism as shown in Fig.~\ref{fig:prototype}(e)-(f), couples radial translation with $90^\circ$ rotation
to provide a wide capture aperture and progressive grasp. A
worm-geared winch provides vertical deployment and recovery while resisting
back driving under load. ArUco-based visual alignment, range sensing, and
state-based control coordinate the alignment-deploy-grasp-recover sequence,
while the mechanically enlarged capture region facilitates tolerating position error positioning error without independent actuation of the six plates.

The main contributions of this work are:
\begin{enumerate}
    \item \textbf{Single-actuator radial grasping mechanism:}
    A six-plate mechanism coupling synchronized radial translation with
    $90^\circ$ plate rotation to provide a wide capture aperture and
    progressive parcel grasp.

    \item \textbf{Winch-deployed autonomous collection:}
    Integration of the gripper with a worm-geared winch, vision and range
    sensing, and state-based control for autonomous
    alignment-deploy-grasp-recover operation.

    \item \textbf{Mechanical and experimental validation:}
Characterization of the proposed mechanism through analytical modeling,
Finite-Element Analysis (FEA), and experimental evaluation of static grasping,
capture workspace, winch performance, and end-to-end outdoor aerial
parcel collection.
\end{enumerate}

The remainder of this letter is organized as follows.
Section~\ref{sec:mechanical_design} presents the mechanical design of the
HexaGripper, followed by the perception and autonomous collection framework
in Section~\ref{sec:perception}. Section~\ref{sec:experiments} evaluates the
proposed system through structural analysis and experimental testing.
The resulting findings are discussed in Section~\ref{sec:discussion}, while
Section~\ref{sec:limitations} presents the limitations and directions for
future work. Finally, Section~\ref{sec:conclusion} concludes the letter.

\section{Mechanical Design}\label{sec:mechanical_design}
This section presents the design architecture and evolution of the
HexaGripper, including the cam and plate mechanisms, actuator sizing,
winch deployment, and prototype fabrication.

\subsection{Design Architecture and Operating Principle}

The HexaGripper comprises six identical plate assemblies arranged
symmetrically around a central rotary cam, as shown in
Fig.~\ref{fig:prototype}. The six extenders are
mechanically synchronized by a single rotary actuator, thereby avoiding
independent actuation of the gripping surfaces. Each extender undergoes 30~mm of radial travel. During the initial 15~mm,
the plates translate radially while approximately maintaining their
orientation. Over the remaining 15~mm, a pin-in-slot mechanism converts the
radial displacement into plate rotation, producing approximately
$90^\circ$ rotation at maximum extension. The fully open configuration
therefore forms a wide, approximately planar capture region, whereas
reversing the cam motion rotates and translates the plates inward to enclose
the parcel. This coupled radial-angular motion reduces the tendency of a
vertically oriented plate to displace an off-center parcel and increases
tolerance to lateral positioning error during pickup. Each gripping plate is 150~mm high and incorporates a 20~mm lead-in region
inclined at $33^\circ$. This geometry guides a slightly misaligned parcel
toward the enclosure during collection while avoiding a sharp $90^\circ$
contact edge. Curved plates were selected such that the six surfaces
approximate a cylindrical profile when closed, providing a compact external
geometry during flight. The aerodynamic benefit of this geometry was not
quantified in the present study.


\subsection{Mechanism Evolution and Selection}

The grasping mechanism was decomposed into three functional requirements:
opening/closing actuation, vertical deployment, and secure parcel retention.
A morphological analysis was then used to systematically generate candidate
configurations from alternative solutions to these functions. Seven grasping
concepts were considered: bird-leg cable actuation~\cite{Askari2024},
independently actuated fingers~\cite{Zhao2025}, a circular-cam
extender~\cite{Jiang2026}, converging struts~\cite{Smith2023}, a
funnel-and-pin interface~\cite{Hoang2024}, a linkage
collector~\cite{Lee2021}, and a geared three-finger
mechanism~\cite{Setty2020}. Each concept was combined with direct and
winch-based deployment~\cite{Kang2025, Lan2024}, resulting in 14 candidate
configurations. The configurations were evaluated using the six weighted criteria in
Table~\ref{tab:weights}. The weighted score was calculated as
$S_j=\sum_{i=1}^{6}w_i s_{ij}$, where $w_i$ is the criterion weight and
$s_{ij}$ is the corresponding score from 1 (poor) to 5 (excellent). The circular-cam mechanism with winch deployment achieved the highest weighted score of 4.05/5 and was selected for further development. The selected mechanism was subsequently refined to improve the capture of
laterally misaligned parcels. The design incorporates $90^\circ$ plate rotation
(Fig.~\ref{fig:prototype}(f)), allowing the plates to approach a
near-horizontal orientation when open and progressively enclose the parcel
during closure. The coupled radial and rotational motion provides a wide
capture region for accommodating lateral parcel-positioning errors.

\begin{table}[t]
\caption{Mechanical design-selection criteria}
\label{tab:weights}
\centering
\small
\begin{tabular}{lc}
\toprule
Criterion & Weight \\
\midrule
Simple actuation                    & 0.15 \\
Low mass                            & 0.15 \\
Position-error tolerance            & 0.20 \\
Repeatability and robustness        & 0.20 \\
Low interference with UAV systems   & 0.15 \\
Package and aircraft safety         & 0.15 \\
\midrule
Total                               & 1.00 \\
\bottomrule
\end{tabular}
\end{table}

\subsection{Archimedean-Spiral Cam Design}
\begin{figure}[t]
    \centering
    \includegraphics[width=0.9\columnwidth]{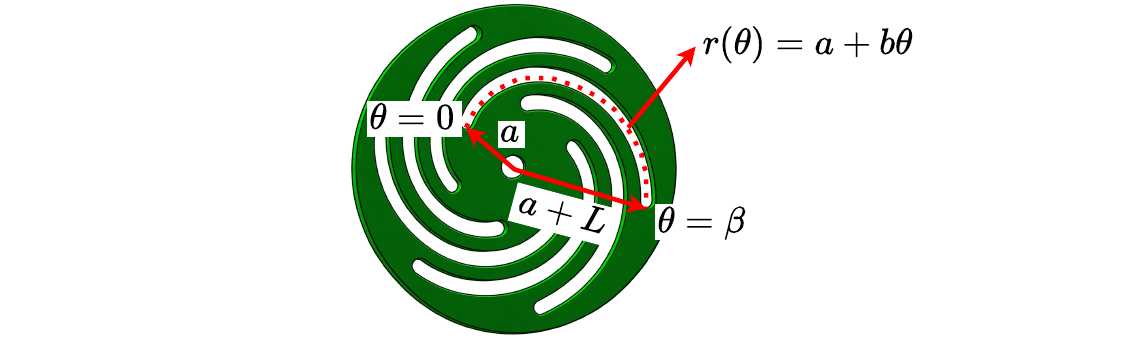}
    \caption{Archimedean-spiral cam geometry showing the initial follower radius
$a$, final radius $a+L$, cam rotation range $0\leq\theta\leq\beta$, and
spiral centerline $r(\theta)=a+b\theta$.}
    \label{fig:arch}
\end{figure}
The six radial extenders are synchronously driven by slots incorporated into
a central rotary cam. An Archimedean-spiral profile~\cite{Liu2021}, defined by
$r(\theta)=a+b\theta$, was selected to provide constant radial displacement
per unit cam rotation, where $r(\theta)$ is the follower radius, $\theta$ is
the cam rotation angle measured from the initial position, $a$ is the initial
follower radius, and $b$ is the radial growth per radian. For a required
radial travel $L$ over a cam rotation $\beta$ as illustrated in Fig.~\ref{fig:arch}, $b=L/\beta$. With $L=30$~mm
and $\beta=150^\circ=2.618$~rad, $b=11.46$~mm/rad. For an ideal radial follower, the instantaneous pressure angle is $\alpha(\theta)=\tan^{-1}[b/r(\theta)]$ and is therefore maximum at the minimum cam radius. With $a=26$~mm, the maximum pressure angle is $\alpha_{\max}=\tan^{-1}(11.46/26)\approx23.8^\circ$, which remains below the adopted $30^\circ$ design limit and decreases as the follower moves outward. For parametric CAD implementation, the spiral centerline was defined as
$x(\theta)=(a+b\theta)\cos(\theta+\theta_0)$ and
$y(\theta)=(a+b\theta)\sin(\theta+\theta_0)$, where $\theta_0$ is the
angular offset used to align the six equally spaced slots with their
followers. The resulting
cam converts a single rotary input into equal 30~mm radial displacements of
all six plate assemblies.

\subsection{Coupled Translational-Rotational Plate Mechanism}
\begin{figure*}[t]
    \centering
    \includegraphics[width=0.95\textwidth]{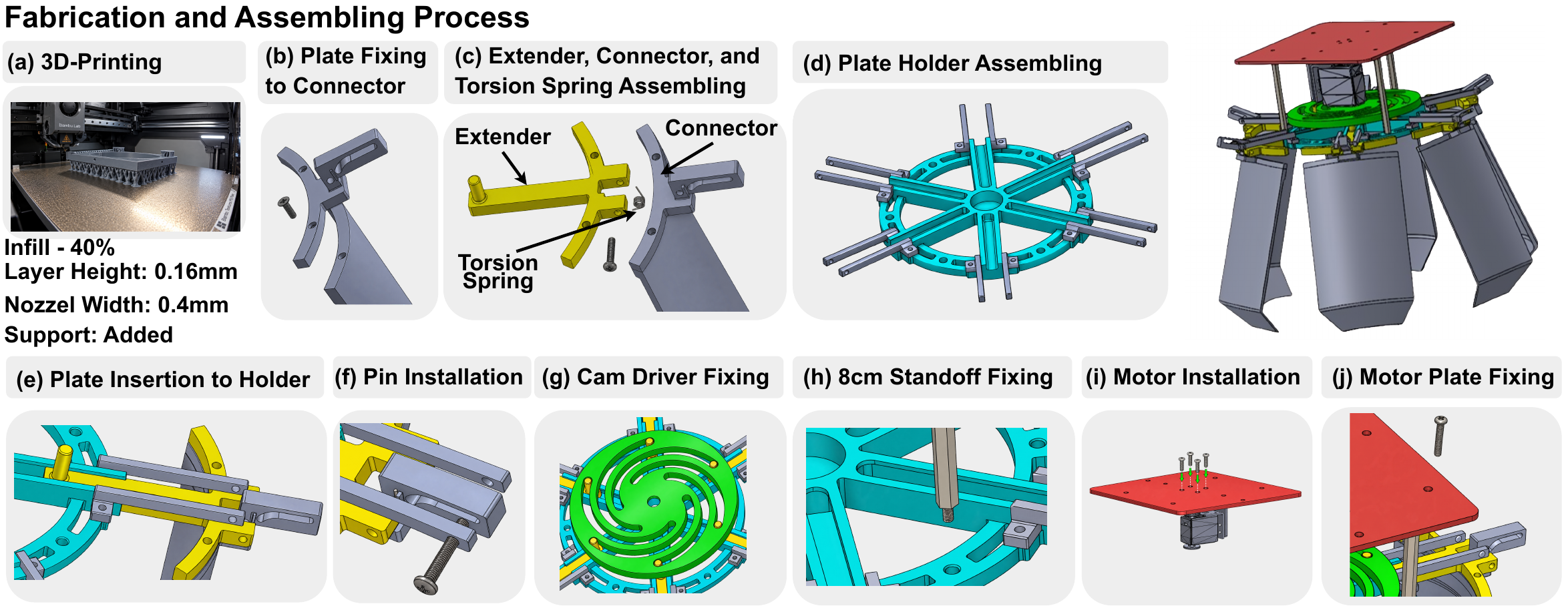}
    \caption{Evolution of the HexaGripper from the selected radial-cam
    concept to the final mechanism. The design progressed from radial-only
    plate translation to coupled radial translation and $90^\circ$ plate
    rotation, with the six plate assemblies synchronized by a single cam
    and deployed using a worm-geared winch.}
    \label{fig:method}
\end{figure*}
\begin{figure}[t]
    \centering
    \includegraphics[width=\columnwidth]{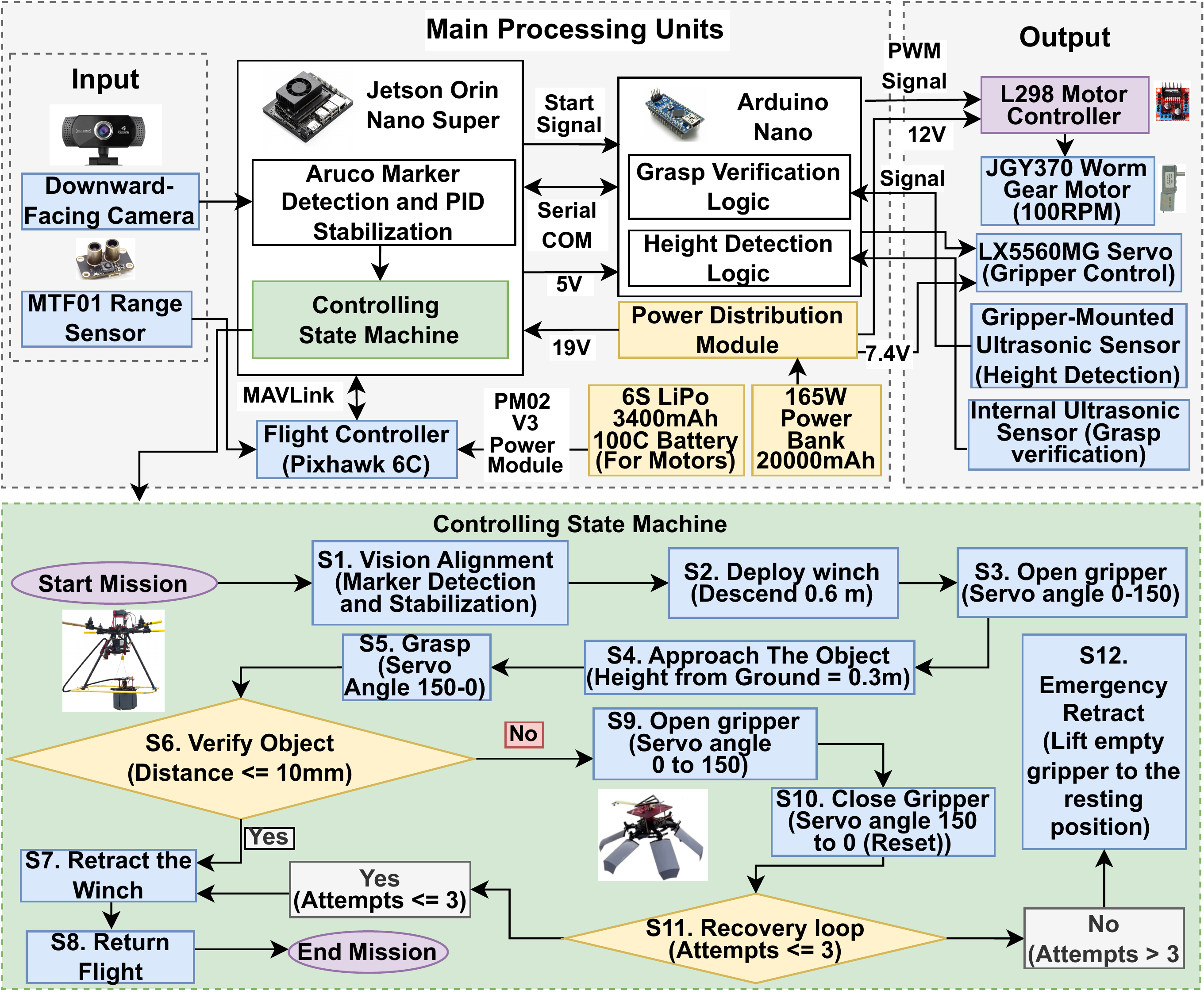}
    \caption{Perception and control architecture with the state-based
    autonomous collection sequence. Visual target localization and range
    sensing guide UAV alignment and gripper deployment, while the controller coordinates winch and gripper actuation, grasp verification,
    and loaded recovery.}
    \label{fig:control}
\end{figure}

Each plate connector was designed to convert the final 15~mm of extender
displacement into angular plate motion, thereby enlarging the capture
aperture without increasing the radial travel of the cam. As shown in
Fig.~\ref{fig:prototype}(d), the connector acts simultaneously as a hinge and a
motion-conversion link. The initial slot geometry consists of a straight
segment followed by a curved section, such that the plate initially
translates radially and subsequently rotates during the final stage of
extension. For a 15~mm arc length producing a nominal $90^\circ$ rotation, the
corresponding slot radius is $R_{\mathrm{slot}}=s/\theta
=15/(\pi/2)=9.55$~mm. In the implemented mechanism, the follower pin is
offset from the plate hinge, and the plate rotation therefore depends on
their relative motion. This relationship is approximated by
$R_l\sin\theta_p=R_{\mathrm{slot}}\sin\phi_s$, where $R_l$ is the distance
between the plate hinge and follower pin, $\theta_p$ is the plate rotation,
and $\phi_s$ is the angular displacement of the pin along the curved slot. The offset geometry amplifies the angular response of the plate, producing
approximately twice the plate rotation relative to the angular displacement
traversed along the slot. Consequently, the required $90^\circ$ plate
rotation is achieved within the available 15~mm radial displacement while
maintaining a compact connector geometry. As illustrated in
Fig.~\ref{fig:prototype}(c) to Fig.~\ref{fig:prototype}(a), this motion transforms the plate from its
closed orientation to the fully open configuration without requiring an
additional actuator.

\subsection{Actuator and Torsion-Spring Design}

The actuator was sized for simultaneous loading of the six cam followers.
The required cam torque~\cite{Cardoso2025} was approximated as
$T_c=NFr\tan(\alpha+\phi)$, where $N=6$ is the number of simultaneously
loaded cam followers, $F$ is the radial force per
follower, $r$ is the instantaneous cam radius, $\alpha$ is the pressure
angle, and $\phi=\tan^{-1}\mu$ is the friction angle. Using an assumed realistic friction coefficient of $\mu=0.45$ during operation, the corresponding
friction angle is $\phi=\tan^{-1}(\mu)\approx24.2^\circ$. This relation was
used for actuator sizing because friction, spring preload, and assembly
effects are difficult to model accurately. An LX-5560MG-class servo with a rated torque of
6.37~N$\cdot$m at 7.4~V was selected. Each plate uses a
torsion spring to resist opening under the parcel load. For a 1~kg design
parcel, assuming equal load distribution among the six plates, the load
supported by each plate is approximately 0.167~kg. Including the 0.050~kg
moving plate assembly, the resulting moment was estimated as
$M_g=(0.167+0.050)(9.81)(0.065)=0.138$~N$\cdot$m. Applying a safety factor
of 1.5 gives a required spring torque of 0.207~N$\cdot$m. With a $10^\circ$ preload and $90^\circ$ plate rotation, the required
torsional spring stiffness was calculated as
$k_s=$ $0.119$~N$\cdot$m/rad. This value was used as the spring-design target.

\subsection{Winch Deployment Mechanism}

The HexaGripper is vertically deployed using a winch to maintain separation
between the UAV and the pickup surface while allowing limited passive
realignment of the suspended gripper. A JGY-370 worm-geared motor was
selected because the transmission resists backdriving and can maintain the
suspended load without continuous motor torque. For a suspended mass
$m_{\mathrm{tot}}=1.2$~kg, effective spool radius
$R_{\mathrm{eff}}=15$~mm, and drivetrain efficiency $\eta=0.65$, the
required output torque was estimated as
$T_w=m_{\mathrm{tot}}gR_{\mathrm{eff}}/\eta=0.271$~N$\cdot$m.
A 100~rpm motor was selected to balance output torque and deployment
speed. The winch uses an 8~mm carbon-steel shaft,
$8\times22\times7$~mm bearings, and an $8\times8$~mm shaft coupler,
and demonstrated a suspended-load capacity of 1.2~kg.

\subsection{Prototype Fabrication and Integration}

The fabrication and assembly process of the HexaGripper is shown in
Fig.~\ref{fig:method}. The gripping plates, holders, radial extenders,
connectors, bearing holders, and winch spool were 3-D printed in PLA+ with
40\% infill, while the gripper base and winch plates were laser-cut from
acrylic. Steel torsion springs, shafts, fasteners, heat-set inserts, and
bearings were used at load-bearing and rotating interfaces. Since the
printed components exhibit orientation-dependent properties, the homogeneous
material approximation used in Sec.~\ref{sec:structural_analysis} is
considered when interpreting the structural results. The integrated gripper
and winch (Fig.~\ref{fig:prototype}(d)) weigh 0.80~kg and 0.52~kg,
respectively, giving a combined mechanical mass of 1.32~kg and leaving
approximately 0.18~kg of the 1.5~kg subsystem allocation for wiring and
interface hardware.

\section{Perception and Autonomous Collection Framework}\label{sec:perception}

\begin{figure}[t]
    \centering
    \includegraphics[width=0.9\columnwidth]{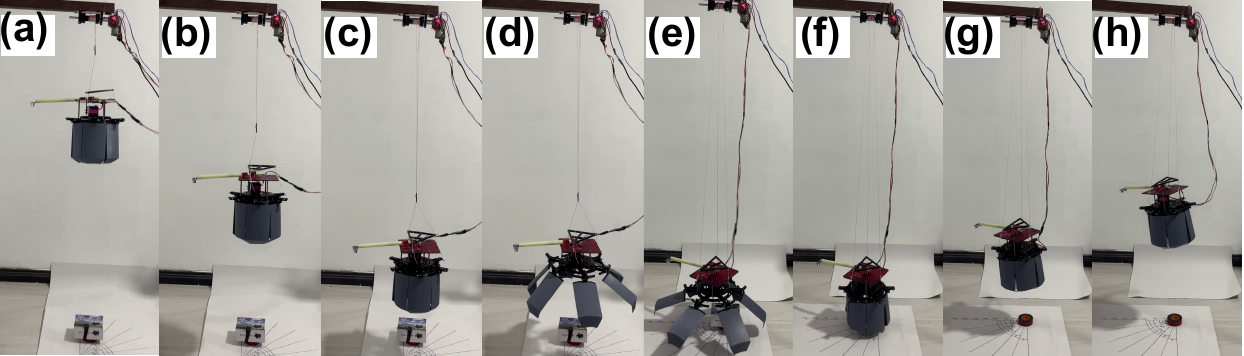}
    \caption{Sequential operation of the winch-deployed gripper during
    parcel collection: (a-c) descent toward the parcel, (d-e) gripper
    opening and engagement, (f) parcel enclosure, and (g-h) winch
    retraction and parcel lifting.}
    \label{fig:gripper_sequence}
\end{figure}

\begin{figure}[t]
    \centering
    \includegraphics[width=0.7\columnwidth]
    {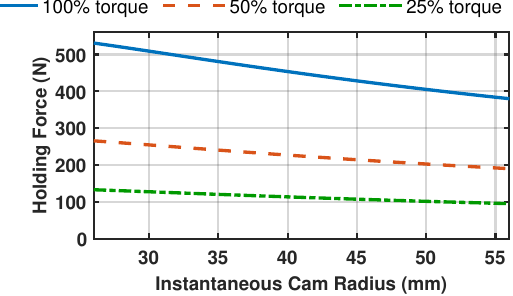}
    \caption{Predicted holding force across the cam radius at different servo torque levels.}
    \label{fig:holding_force}
\end{figure}

\begin{figure}[t]
    \centering
    \includegraphics[width=0.75\columnwidth]{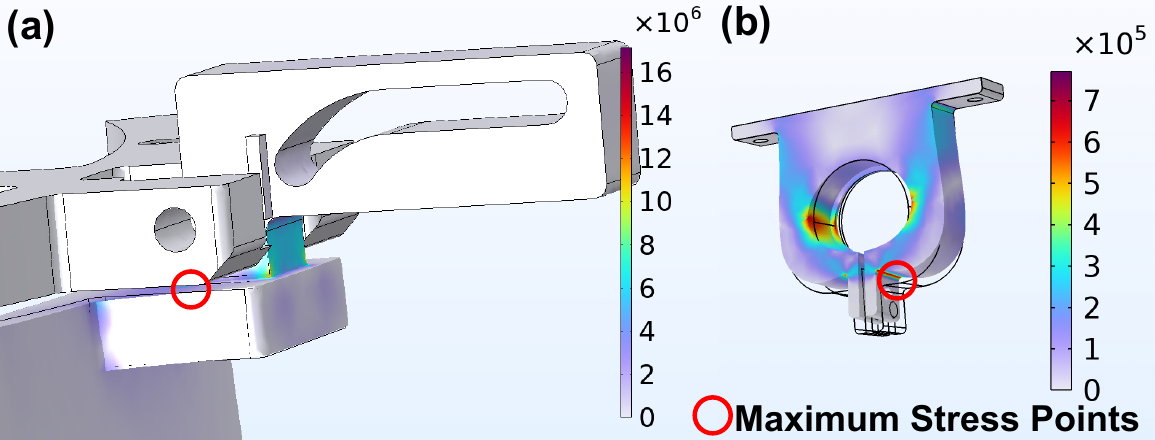}
    \caption{Finite-element stress analysis of
    (a) the gripper plate assembly under a 10~N gripping load and
    (b) the winch bearing bracket under a 30~N vertical load.}
    \label{fig:fea}
\end{figure}

\begin{figure}[t]
    \centering
    \begin{minipage}[t]{0.45\columnwidth}
        \centering
        \includegraphics[width=\linewidth]
        {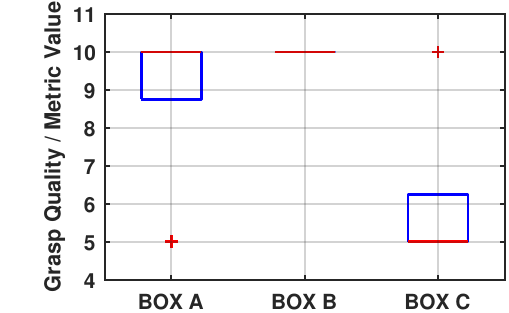}\\[-1mm]
        {\footnotesize (a) Grasp-quality repeatability}
    \end{minipage}%
    \hspace{0.08\columnwidth}%
    \begin{minipage}[t]{0.4\columnwidth}
        \centering
        \includegraphics[width=\linewidth]
        {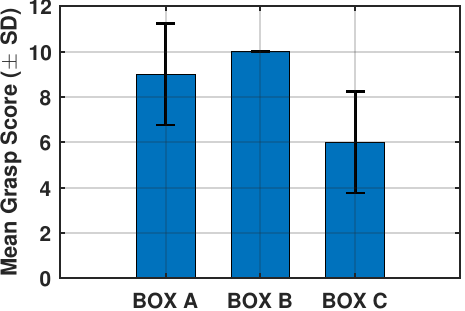}\\[-1mm]
        {\footnotesize (b) Mean grasp-quality score}
    \end{minipage}
    \caption{Static grasp performance across the three parcel geometries:
    (a) trial-to-trial grasp-quality repeatability and
    (b) mean grasp-quality score with standard-deviation error bars.}
    \label{fig:staticplots}
\end{figure}

The autonomous collection framework integrates a downward-facing camera,
ultrasonic range sensor, winch, and gripper servo with the UAV flight
controller, as shown in Fig.~\ref{fig:control}. ArUco-based localization,
following the method described in~\cite{Adikaram2026}, provides lateral
alignment, while ultrasonic ranging determines the final gripper-to-parcel
distance. The state-based sequence performs marker search and alignment,
winch descent, range verification, gripper opening, final descent, closure,
grasp verification, and loaded ascent. Failed grasp verification triggers a
retry. Lateral positioning error is accommodated by the gripper's
capture aperture, reducing the need for full six-degree-of-freedom parcel-pose
estimation.

\section{Experimental Evaluation}\label{sec:experiments}

This section evaluates the HexaGripper through grasp-force modeling, FEA,
static grasping, capture-workspace, winch-performance, and integrated
aerial collection tests. The complete parcel-collection sequence under the
static experimental setup is illustrated in Fig.~\ref{fig:gripper_sequence}.

\subsection{Grasp-Force Modeling}
\label{sec:structural_analysis}
A quasi-static model estimated the radial cam-follower force as
$F_{\mathrm{cam}}=T_{\mathrm{mot}}/[R_{\mathrm{inst}}\tan(\alpha+\phi)]$,
with $R_{\mathrm{inst}}=26$-$56$~mm. The plate force is
$F_{\mathrm{grip}}=0.892F_{\mathrm{cam}}$, giving
$F_{\mathrm{hold}}=6\mu_pF_{\mathrm{grip}}$. At the specified servo torque
of 6.37~N$\cdot$m at 7.4~V, the predicted holding force decreased from
531 to 380~N across the cam travel, corresponding to equivalent supported
masses of 54.1-38.7~kg as shown in Fig.~\ref{fig:holding_force}. At 50\% torque, the maximum equivalent supported
mass was 27.1~kg. These values are theoretical force-transmission limits
rather than UAV payload ratings and neglect dynamic and contact effects.

\subsection{Finite-Element Analysis}

Stationary FEA were performed in COMSOL for the gripper
plate assembly and winch bearing bracket. The printed components were modeled
as homogeneous, isotropic PLA+ with density 496~kg/m$^3$, Poisson's ratio
0.39, and Young's modulus 2.96~GPa. A 1-10~N load sweep on the plate
assembly produced a maximum von Mises stress of 16.67~MPa at 10~N near the
connector-hinge region (Fig.~\ref{fig:fea}(a)), corresponding to a nominal
FOS of 2.10 relative to an assumed 35~MPa tensile strength. A 5-30~N sweep
on the winch bracket produced 1.10~MPa at 30~N near the bearing support
(Fig.~\ref{fig:fea}(b)), giving an FOS greater than 30. Thus, the
connector-hinge region was the more critical modeled location. These FOS
values are nominal estimates because the homogeneous isotropic model neglects
print anisotropy, voids, fastener effects, and cyclic loading.

\subsection{Static Grasp Performance} 

Static experiments evaluated grasp repeatability across three cardboard
parcels: Box A ($12.5\times10\times5.5$~cm), Box B
($16\times14.5\times6$~cm), and Box C ($11\times9\times8$~cm). The
gripper was hanged on a stationary test apparatus as shown in Fig.~\ref{fig:gripper_sequence} and centered above each parcel, with five repeated trials performed for each geometry. Grasp quality was scored as 10 for complete grasp, 5 for partial grasp, and 0 for a failed grasp. As shown in Fig.~\ref{fig:staticplots} and summarized in Table~\ref{tab:grasp_results}, all 15 trials resulted in successful pickup, corresponding to a 100\% pickup success rate. Of these, 10/15 achieved complete grasp. Box B achieved complete grasp in all five trials, while Boxes A and C achieved mean grasp-quality scores of 9.0 and 6.0, respectively. The overall mean grasp-quality score was 8.33/10. The variation in grasp quality indicates sensitivity to parcel geometry despite the 100\% pickup success under centered conditions. The larger horizontal footprint of Box B resulted in more consistent multi-plate engagement, whereas the smaller Box C more frequently resulted in partial grasp.

\begin{table}[t]
\caption{Static and workspace grasp-quality results}
\label{tab:grasp_results}
\centering
\small
\setlength{\tabcolsep}{4pt}

\begin{tabular}{cccccccc}
\toprule
\multicolumn{8}{c}{\textbf{Static Grasp Results}}\\
\midrule
Box & $L$ & $W$ & $H$ & \multicolumn{3}{c}{Scores} & Mean\\
    & (cm) & (cm) & (cm) & \multicolumn{3}{c}{(Trial 1 to 5)} & \\
\midrule
A & 12.5 & 10.0 & 5.5 & \multicolumn{3}{c}{10, 10, 5, 10, 10} & 9.0\\
B & 16.0 & 14.5 & 6.0 & \multicolumn{3}{c}{10, 10, 10, 10, 10} & 10.0\\
C & 11.0 & 9.0 & 8.0 & \multicolumn{3}{c}{5, 5, 5, 10, 5} & 6.0\\

\midrule
\multicolumn{8}{c}{\textbf{Workspace Grasp-Quality Results}}\\
\midrule
Radius & \multicolumn{7}{c}{Target angle (deg)}\\
(mm) & 0 & 15 & 30 & 45 & 60 & 75 & 90\\
\midrule
0   & 10 & 10 & 10 & 10 & 10 & 10 & 10\\
50  & 10 & 10 & 10 & 10 & 10 & 10 & 10\\
100 & 10 & 10 & 5  & 5  & 5  & 10 & 10\\
150 & 5 & 5  & 10  & 10  & 10  & 5  & 10\\
200 & 0  & 0  & 0  & 0  & 0  & 0  & 0\\
\bottomrule
\end{tabular}

\vspace{2pt}
\centering
\footnotesize
\textit{Score: 10 = successful grasp,\quad 5 = partial grasp,\quad 0 = failed grasp.}

\end{table}

\begin{table}[t]
\caption{Real-world aerial collection results during UAV hovering.}
\label{tab:dynamic}
\centering
\small
\setlength{\tabcolsep}{5pt}
\renewcommand{\arraystretch}{1.15}

\begin{tabular}{c c c c c c}
\toprule
\textbf{Trial} & \textbf{Alignment} & \textbf{Deploy} &
\textbf{Grasp} & \textbf{Recover} & \textbf{Overall} \\
\midrule

1 & \cmark & \cmark & \cmark & \cmark & \cmark \\
2 & \cmark & \cmark & \cmark & \cmark & \cmark \\
3 & \cmark & \cmark & \xmark & -      & \xmark \\
4 & \cmark & \cmark & \cmark & \cmark & \cmark \\
5 & \cmark & \cmark & \cmark & \cmark & \cmark \\

\bottomrule
\end{tabular}

\vspace{2pt}
\footnotesize
\cmark~= successful stage,\quad
\xmark~= failed stage,\quad
- = not attempted.

\end{table}

\begin{table*}[ht]
\centering
\caption{Comparative evaluation of representative aerial grasping and perching mechanisms.}
\label{tab:gripper_comparison}
\footnotesize
\setlength{\tabcolsep}{3.5pt}
\renewcommand{\arraystretch}{1.12}

\begin{tabular}{lclcccccccc}
\toprule
\begin{tabular}[c]{@{}c@{}}Prior\\Works\end{tabular} &
Year &
Application &
\begin{tabular}[c]{@{}c@{}}Single\\Act.?\end{tabular} &
\begin{tabular}[c]{@{}c@{}}Tethered\\Deploy.?\end{tabular} &
\begin{tabular}[c]{@{}c@{}}Enclosure\\Grasp?\end{tabular} &
\begin{tabular}[c]{@{}c@{}}Position\\Tolerant?\end{tabular} &
\begin{tabular}[c]{@{}c@{}}Contact\\Elements\end{tabular} &
\begin{tabular}[c]{@{}c@{}}Gripper\\Mass (g)\end{tabular} &
\begin{tabular}[c]{@{}c@{}}Capture\\Offset (mm)\end{tabular} &
\begin{tabular}[c]{@{}c@{}}Aerial\\Success (\%)\end{tabular} \\
\midrule


RAPTOR~\cite{Appius2022}
& 2022 & Rapid aerial pickup
& \xmark & \xmark & \xmark & \cmark
& 4 & - & - & 83 \\


Linkage Gripper~\cite{Smith2023}
& 2023 & Object pick-and-place
& \cmark & \xmark & \xmark & \xmark
& 2 & - & - & 55 \\

TRIGGER~\cite{Kremer2023}
& 2023 & Universal jamming
& \xmark & \xmark & \cmark & \cmark
& 1 & 380 & - & - \\

Modular Soft SAV~\cite{Cheung2024}
& 2024 & Grasping \& landing
& \cmark & \xmark & \cmark & \cmark
& 4 & 260 & - & - \\

Multitentacle~\cite{Yang2024}
& 2024 & Dynamic capture
& \xmark & \xmark & \cmark & \cmark
& 12 & - & - & - \\


Passive Dynamic~\cite{Firouzeh2024}
& 2024 & Dynamic perching
& Passive & \xmark & \cmark & \cmark
& 2 & 150 & 40 & - \\


Bistable Soft Net~\cite{Yang2025}
& 2025 & Grasp \& perch
& \cmark & \xmark & \cmark & \cmark
& 4 & 197 & - & - \\

SPIBOT~\cite{Kang2025}
& 2025 & Cargo handling
& \xmark & \cmark & \cmark & \cmark
& 6 & - & - & - \\

SPIDAR~\cite{Zhao2025}
& 2025 & 3D multi-limb grasp
& \xmark & \xmark & \cmark & \xmark
& 4 & - & - & - \\

Swooper~\cite{Huang2026}
& 2026 & High-speed swooping
& \cmark & \xmark & \xmark & \xmark
& 2 & - & 50 & 84 \\

Cable Differential~\cite{Jiang2026}
& 2026 & Fixed-position grasp
& \cmark & \xmark & \xmark & \cmark
& 2 & 490 & 4.02 & - \\

\midrule
\textbf{Ours (HexaGripper)}
& \textbf{2026}
& \textbf{Parcel collection}
& \cmark
& \cmark
& \cmark
& \cmark
& \textbf{6}
& \textbf{800}
& \textbf{150}
& \textbf{80} \\

\bottomrule
\end{tabular}

\vspace{1.5mm}
\begin{minipage}{0.98\textwidth}
\scriptsize
\cmark~and \xmark~denote presence and absence of the corresponding characteristic, respectively; ``-'' denotes unstated, non-reported, or non-comparable experimental data. \textit{Passive} indicates mechanically impact-triggered closure without motor actuation during impact.
\end{minipage}
\end{table*}

\begin{figure}[t]
    \centering
    \begin{minipage}[t]{0.45\columnwidth}
        \centering
        \includegraphics[width=\linewidth]{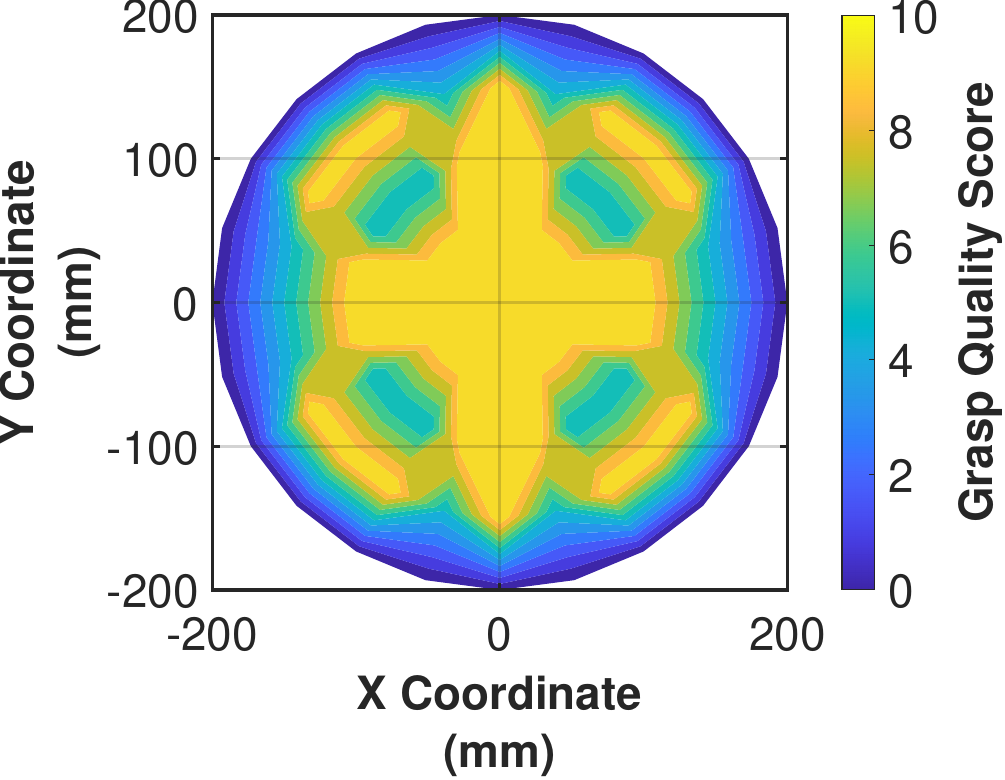}\\[-1mm]
        {\footnotesize (a) Measured 2-D grasp-quality map}
    \end{minipage}
    \hspace{0.06\columnwidth}%
    \begin{minipage}[t]{0.47\columnwidth}
        \centering
        \includegraphics[width=\linewidth]{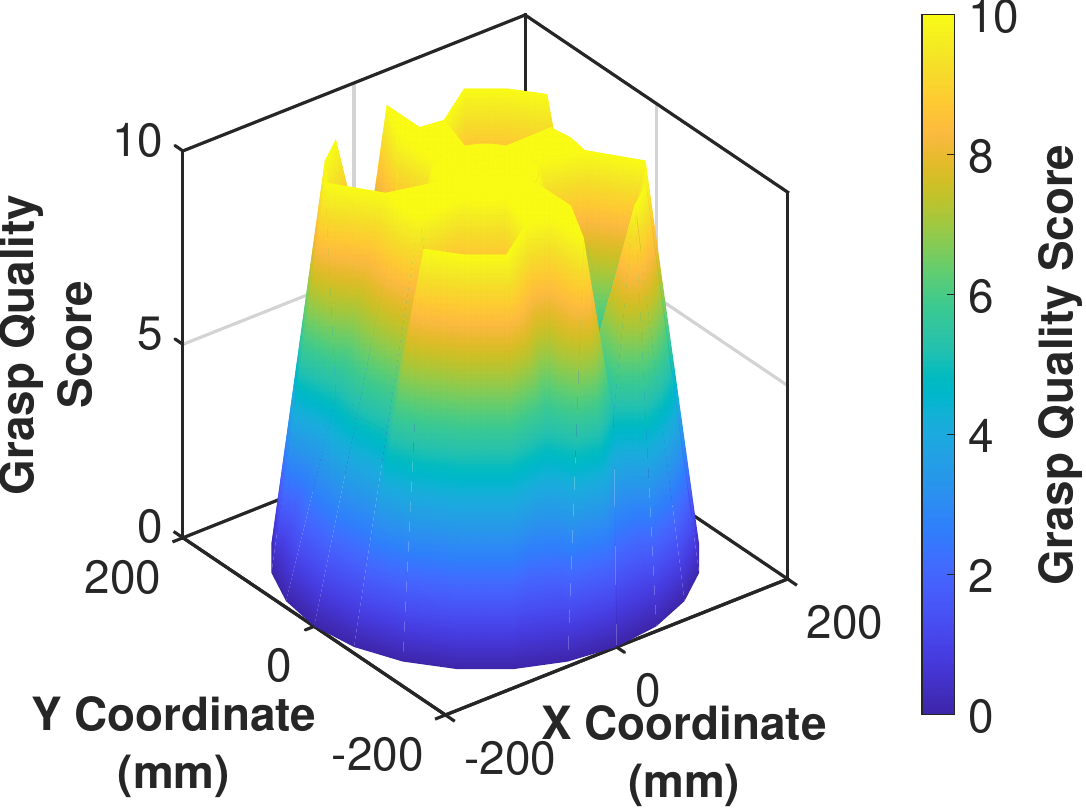}\\[-1mm]
        {\footnotesize (b) Corresponding 3-D score surface}
    \end{minipage}
    \caption{Grasp-quality workspace characterization: (a) measured 2-D grasp-quality map and (b) corresponding 3-D score surface. The experimentally demonstrated reliable grasp radius was 150~mm.}
    \label{fig:workspace}
\end{figure}

\begin{figure}[t] \centering \includegraphics[width=0.75\columnwidth]{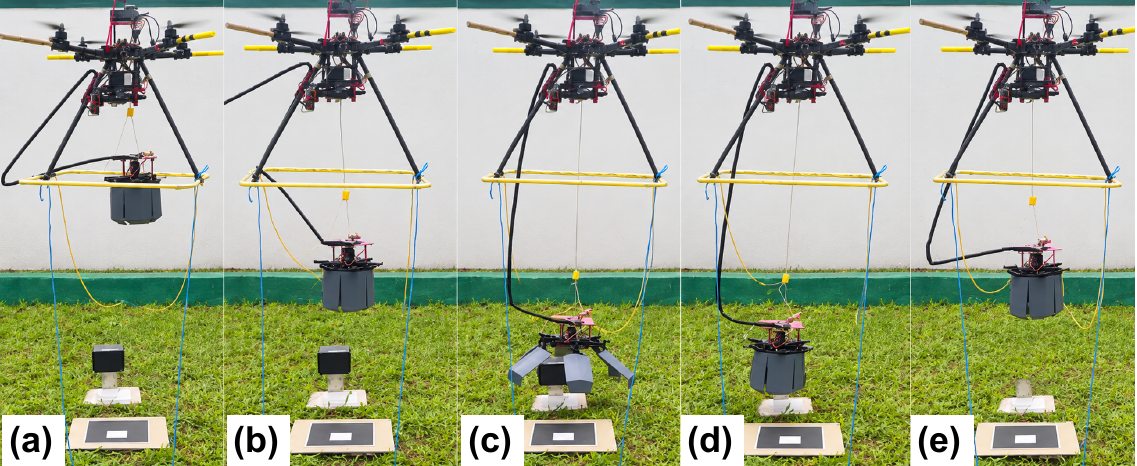} \caption{Outdoor aerial collection sequence showing target alignment, gripper deployment, parcel grasp, and loaded recovery.} \label{fig:dynamic} \end{figure}



\subsection{Capture-Workspace Evaluation}

The lateral capture workspace was evaluated using the 100~g Box B with the
UAV suspended approximately 1.2~m above the test surface. The parcel was
positioned at radial offsets of 0, 50, 100, 150, and 200~mm from the gripper
center. For each radius, one quadrant was sampled at $15^\circ$ angular
increments from $0^\circ$ to $90^\circ$, with symmetry used to represent the
remaining directions. One grasp was performed at each tested coordinate using
the same 0/5/10 grasp-quality metric. As summarized in
Table~\ref{tab:grasp_results} and Fig.~\ref{fig:workspace}, successful pickup
was achieved at all 28 tested configurations within radial offsets of
0-150~mm, whereas all seven coordinates tested at 200~mm resulted in failed
grasps. Of the 28 successful grasps, 22 achieved complete enclosure, with a
mean grasp-quality score of 8.93/10. A demonstrated capture radius of 150~mm
was therefore adopted, corresponding to a capture area of
$A_{\mathrm{cap}}=\pi(150)^2=70{,}686$~mm$^2$. This radius is approximately
54.5\% of the nominal 275~mm geometric reach, indicating that geometric
overlap alone does not guarantee sufficient plate-parcel engagement. Since
only one trial was conducted at each coordinate, the measured region
represents an empirical capture boundary rather than a probabilistic success
map.


\subsection{Winch Performance}

Winch performance was evaluated over a vertical travel distance of 1.05~m,
with three repetitions performed for unloaded gripper descent and loaded
ascent. The mean descent velocity was 4.87~cm/s, corresponding to an average
travel time of approximately 21.6~s, while loaded ascent achieved
4.16~cm/s with an average travel time of approximately 25.2~s. The suspended
load therefore reduced the mean winch velocity by approximately 14.6\%. The loaded ascent required a longer travel time than unloaded descent and therefore constituted the dominant component of the winch deployment-recovery cycle. These measurements characterize the period during which the UAV must maintain stable hover and do not include the preceding vision-based target-search and alignment stages.


\subsection{Integrated Aerial Collection}

The complete system was evaluated through five outdoor aerial collection
trials with the UAV hovering above the pickup region, as illustrated in
Fig.~\ref{fig:dynamic}. Each trial comprised four evaluated stages:
ArUco-based alignment, winch deployment, grasp execution, and loaded
recovery. A trial was considered successful only when all four stages were
completed. As summarized in Table~\ref{tab:dynamic}, ArUco-based target alignment and winch
deployment succeeded in all five trials. Four trials completed the full
alignment-deploy-grasp-recover sequence, corresponding to an overall
collection success rate of 80\%. Trial 3 failed during grasp execution when
a sudden gust disturbed the relative position of the suspended gripper and
parcel, causing the parcel to slip before secure grasp. Loaded recovery
was therefore not attempted in this trial. The four successful trials demonstrate the ability of the capture region to
accommodate positioning error during integrated aerial operation.
However, the gust-induced failure indicates that the static capture
workspace does not directly represent dynamic robustness, as relative
lateral motion during closure can reduce plate-parcel overlap and prevent
secure grasp.

\section{Discussion}\label{sec:discussion}

\subsection{Capture Tolerance and Structural Performance}

The HexaGripper synchronizes six contact surfaces using a single actuator,
while its coupled radial-angular motion provides tolerance to lateral
positioning errors. Despite a nominal open radius of 275~mm, reliable
grasping was demonstrated only up to a 150~mm offset, indicating that
capture also depends on plate engagement, contact geometry, and friction for the tested cuboid parcel geometries. The outdoor failure further highlights the effects of tether motion and wind
disturbances. The gripper and winch have a combined mass of 1.32~kg,
making mass a key design constraint, while practical payload is limited by
UAV lift, the 1.5~kg winch capacity, structural strength, servo limits, and
allowable contact force. FEA identified the plate connector-hinge region
as the critical modeled component, reaching 16.67~MPa under a 10~N load,
compared with 1.10~MPa for the winch bracket under 30~N, subject to the
homogeneous isotropic approximation of the printed PLA+ components.

\subsection{Comparison With Prior Aerial Grippers}

Table~\ref{tab:gripper_comparison} highlights the design tradeoffs among
representative aerial grasping and perching mechanisms. Several systems
provide grasp-based and position-tolerant grasping
\cite{Lee2021,Hsiao2022,Cheung2024,Yang2025}, while others prioritize
lightweight passive operation~\cite{Askari2024,Firouzeh2024,Lee2025} or
rapid aerial capture~\cite{Huang2026,Appius2022}. Tethered deployment is
less common, with SPIBOT~\cite{Kang2025} providing this capability using
multiple actuators. In comparison, HexaGripper combines single-actuator
operation, tethered deployment, grasping, and position tolerance.
It demonstrated a 150~mm capture offset and an 80\% outdoor aerial collection success
rate. Its 800~g mass results from the assembly of components
required to provide synchronized six-plate grasping for parcel collection.

\section{Limitations and Future Work}\label{sec:limitations}

The evaluation was limited to three cardboard parcel geometries, one trial
per workspace configurations, and five outdoor collection trials. Parcel mass,
surface friction, compliance, wind velocity, and
tether oscillation were not systematically varied. Furthermore, the FEA
assumed homogeneous isotropic properties for the printed components and did
not consider interlayer anisotropy or fatigue. Future work will therefore
evaluate a broader range of parcel properties and controlled disturbances,
perform repeated workspace and structural tests, reduce subsystem mass,
introduce compliant high-friction contact surfaces, improve tether
stabilization, and evaluate autonomous parcel release.

\section{Conclusion}\label{sec:conclusion}

This work presented HexaGripper, a single-actuator, winch-deployed
six-plate gripper for autonomous aerial parcel collection. The mechanism
combines synchronized radial translation and $90^\circ$ plate rotation
for progressive parcel grasp. Integration with a non-backdrivable
worm-geared winch, visual localization, range sensing, and state-based
control enables autonomous alignment-deploy-grasp-recover operation.

Experimental evaluation achieved 100\% static pickup success
(15/15 trials) across three parcel geometries. All 28 tested positions
within a 150~mm radial offset were successfully grasped, demonstrating
a capture area of 70{,}686~mm$^2$. FEA predicted
maximum stresses of 16.67~MPa in the plate assembly and 1.10~MPa
in the winch bearing bracket. Outdoor aerial collection achieved
80\% success (4/5 trials). These results demonstrate the feasibility
of capture-tolerant aerial parcel collection using mechanically
synchronized grasp and winch-based deployment.

\bibliographystyle{IEEEtran}
\bibliography{Ref}

@ARTICLE{Firouzeh2024,
  author={Firouzeh, Amir and Lee, Jongeun and Yang, Hyunsoo and Lee, Dongjun and Cho, Kyu-Jin},
  journal={IEEE Transactions on Robotics}, 
  title={Perching and Grasping Using a Passive Dynamic Bioinspired Gripper}, 
  year={2024},
  volume={40},
  number={},
  pages={213-225},
  doi={10.1109/TRO.2023.3336216}}

@article{Setty2020,
title = {Generic gripper for an unmanned aerial vehicle},
journal = {Procedia CIRP},
volume = {91},
pages = {486-488},
year = {2020},
note = {Enhancing design through the 4th Industrial Revolution Thinking},
issn = {2212-8271},
doi = {https://doi.org/10.1016/j.procir.2020.02.203},
author = {Kiran Setty and Theo {van Niekerk} and Riaan Stopforth}
}

@ARTICLE{Yang2025,
  author={Yang, Yi and Fan, Linfeng and Weng, Tao and Zhao, Yi and Chen, Bingxing and Li, Wenqiang},
  journal={IEEE Robotics and Automation Letters}, 
  title={Bistable Soft Gripper With Tension Net Applied to UAV}, 
  year={2025},
  volume={10},
  number={2},
  pages={1920-1927},
  doi={10.1109/LRA.2025.3526570}}

@INPROCEEDINGS{Appius2022,
  author={Appius, Aurel X. and Bauer, Erik and Blöchlinger, Marc and Kalra, Aashi and Oberson, Robin and Raayatsanati, Arman and Strauch, Pascal and Suresh, Sarath and von Salis, Marco and Katzschmann, Robert K.},
  booktitle={2022 IEEE/RSJ International Conference on Intelligent Robots and Systems (IROS)}, 
  title={RAPTOR: Rapid Aerial Pickup and Transport of Objects by Robots}, 
  year={2022},
  volume={},
  number={},
  pages={349-355},
  doi={10.1109/IROS47612.2022.9981668}}

@INPROCEEDINGS{Kang2025,
  author={Kang, Gyuree and Güneş, Ozan and Lee, Seungwook and Azhari, Maulana Bisyir and Shim, David Hyunchul},
  booktitle={2025 IEEE International Conference on Robotics and Automation (ICRA)}, 
  title={SPIBOT: A Drone-Tethered Mobile Gripper for Robust Aerial Object Retrieval in Dynamic Environments}, 
  year={2025},
  volume={},
  number={},
  pages={7533-7538},
  doi={10.1109/ICRA55743.2025.11128830}}

@ARTICLE{Yang2024,
  author={Yang, Chenghao and Walker, Ian D. and Branson, David T. and Dai, Jian S. and Sun, Tao and Kang, Rongjie},
  journal={IEEE Transactions on Robotics}, 
  title={A Multitentacle Gripper for Dynamic Capture}, 
  year={2024},
  volume={40},
  number={},
  pages={4284-4300},
  doi={10.1109/TRO.2024.3454437}}

@INPROCEEDINGS{Smith2023,
  author={Smith, Sean and Buchanan, Scott and Pan, Ya-Jun},
  booktitle={2023 IEEE/ASME International Conference on Advanced Intelligent Mechatronics (AIM)}, 
  title={A Linkage-Based Gripper Design with Optimized Data Transmission for Aerial Pick-and-Place Tasks}, 
  year={2023},
  volume={},
  number={},
  pages={446-451},
  doi={10.1109/AIM46323.2023.10196128}}

@INPROCEEDINGS{Lan2024,
  author={Lan, Tian and Romanello, Luca and Kovac, Mirko and Armanini, Sophie F. and Bahadir Kocer, Basaran},
  booktitle={2024 IEEE International Conference on Robotics and Automation (ICRA)}, 
  title={Aerial Tensile Perching and Disentangling Mechanism for Long-Term Environmental Monitoring}, 
  year={2024},
  volume={},
  number={},
  pages={3827-3833},
  doi={10.1109/ICRA57147.2024.10609975}}

@article{Liu2021,
  author  = {Liu, Y. and Cui, S. and Sun, Y.},
  title   = {Mechanical design and analysis of a novel variable stiffness actuator with symmetrical pivot adjustment},
  journal = {Frontiers of Mechanical Engineering},
  volume  = {16},
  pages   = {711--725},
  year    = {2021},
  doi     = {10.1007/s11465-021-0647-1}
}

@article{Zieher2024,
author = {Simon Zieher and Ertug Olcay and Klaus Kefferpütz and Babak Salamat and Sebastian Olzem and Gerhard Elsbacher and Henri Meeß},
title = {Drones for automated parcel delivery: Use case identification and derivation of technical requirements},
journal = {Transportation Research Interdisciplinary Perspectives},
volume = {28},
pages = {101253},
year = {2024},
issn = {2590-1982},
doi = {https://doi.org/10.1016/j.trip.2024.101253}
}

@ARTICLE{Kremer2023,
  author={Kremer, Paul and Nohooji, Hamed Rahimi and Sanchez-Lopez, Jose Luis and Voos, Holger},
  journal={IEEE Access}, 
  title={TRIGGER: A Lightweight Universal Jamming Gripper for Aerial Grasping}, 
  year={2023},
  volume={11},
  number={},
  pages={50098-50115},
  doi={10.1109/ACCESS.2023.3276486}}

@ARTICLE{Lee2025,
  author={Lee, Adel and Shenghai, Yuan and Shen, Zhefu and Zhang, Hongying},
  journal={IEEE Robotics and Automation Letters}, 
  title={MagPerch: A Magnetically Locked, Lightweight Perching Mechanism for Micro-Aerial-Vehicles}, 
  year={2025},
  volume={10},
  number={10},
  pages={10370-10377},
  doi={10.1109/LRA.2025.3597896}}

@ARTICLE{Huang2026,
  author={Huang, Ziken and Niu, Xinze and Chai, Bowen and Jin, Renbiao and Zou, Danping},
  journal={IEEE Robotics and Automation Letters}, 
  title={Swooper: Learning High-Speed Aerial Grasping With a Simple Gripper}, 
  year={2026},
  volume={11},
  number={2},
  pages={2298-2305},
  doi={10.1109/LRA.2025.3643298}}

@Article{Cardoso2025,
AUTHOR = {Cardoso, Daniel Silva and Fael, Paulo Oliveira and Gaspar, Pedro Dinis and Espírito-Santo, António},
TITLE = {Balancing Cam Mechanism for Instantaneous Torque and Velocity Stabilization in Internal Combustion Engines: Simulation and Experimental Validation},
JOURNAL = {Energies},
VOLUME = {18},
YEAR = {2025},
NUMBER = {13},
ARTICLE-NUMBER = {3256},
URL = {https://www.mdpi.com/1996-1073/18/13/3256},
ISSN = {1996-1073},
DOI = {10.3390/en18133256}
}

@ARTICLE{Porghoveh2026,
  author={Porghoveh, Mojtaba and Carli, Raffaele and Dotoli, Mariagrazia},
  journal={IEEE Access}, 
  title={Drones in Automated and Robotized Internal Logistics: A Review of Optimization and Control Techniques}, 
  year={2026},
  volume={14},
  number={},
  pages={76138-76172},
  doi={10.1109/ACCESS.2026.3692765}}

@INPROCEEDINGS{Domislovic2025,
  author={Domislovic, Jakob and Milijas, Robert and Ivanovic, Antun and Car, Marko and Vasiljevic, Goran and Ferreira, Barbara Arbanas and Petric, Frano and Orsag, Matko and Bogdan, Stjepan},
  booktitle={2025 International Conference on Unmanned Aircraft Systems (ICUAS)}, 
  title={$\rho$LiRLo: LiDAR-Based Relative Localization with Retro-Reflective Marker}, 
  year={2025},
  volume={},
  number={},
  pages={228-235},
  doi={10.1109/ICUAS65942.2025.11007861}}

@INPROCEEDINGS{Cheung2024,
  author={Cheung, Hiu Ching and Chang, Ching-Wei and Jiang, Bailun and Wen, Chih-Yung and Chu, Henry K.},
  booktitle={2024 IEEE 7th International Conference on Soft Robotics (RoboSoft)}, 
  title={A Modular Pneumatic Soft Gripper Design for Aerial Grasping and Landing}, 
  year={2024},
  volume={},
  number={},
  pages={82-88},
  doi={10.1109/RoboSoft60065.2024.10521918}}

@ARTICLE{Hsiao2022,
  author={Hsiao, HaoTse and Sun, Jiefeng and Zhang, Haijie and Zhao, Jianguo},
  journal={IEEE/ASME Transactions on Mechatronics}, 
  title={A Mechanically Intelligent and Passive Gripper for Aerial Perching and Grasping}, 
  year={2022},
  volume={27},
  number={6},
  pages={5243-5253},
  doi={10.1109/TMECH.2022.3175649}}

@INPROCEEDINGS{Zhao2025,
  author={Zhao, Moju},
  booktitle={2025 IEEE International Conference on Robotics and Automation (ICRA)}, 
  title={Aerial Grasping by Multi-Limbed Flying Robot SPIDAR Based on Vectored Thrust Control}, 
  year={2025},
  volume={},
  number={},
  pages={5202-5208},
  doi={10.1109/ICRA55743.2025.11128075}}

@ARTICLE{Nishimura2026,
  author={Nishimura, Toshihiro and Okatsu, Hono and Watanabe, Tetsuyou},
  journal={IEEE Robotics and Automation Letters}, 
  title={Single-Motor-Driven Robotic Hand With Multiple Grippers Using Gravity-Driven Torque-Path Switching Mechanism}, 
  year={2026},
  volume={11},
  number={6},
  pages={7246-7253},
  doi={10.1109/LRA.2026.3685935}}

@INPROCEEDINGS{Wang2025,
  author={Wang, Yongyue and Zhu, Mingzhu and Yue, Xiaokui},
  booktitle={2025 IEEE International Conference on Real-time Computing and Robotics (RCAR)}, 
  title={Bistable Miura-ori soft gripper for adaptive and rapid capture*}, 
  year={2025},
  volume={},
  number={},
  pages={781-786},
  doi={10.1109/RCAR65431.2025.11139530}}

@ARTICLE{Askari2024,
  author={Askari, Mohammad and Shin, Won Dong and Lenherr, Damian and Stewart, William and Floreano, Dario},
  journal={IEEE/ASME Transactions on Mechatronics}, 
  title={Avian-Inspired Claws Enable Robot Perching or Walking}, 
  year={2024},
  volume={29},
  number={3},
  pages={1856-1866},
  doi={10.1109/TMECH.2023.3331357}}

@ARTICLE{Jiang2026,
  author={Jiang, Xiao and Shen, Jiahao and Huang, Diwei and Cheng, Xiang and Xia, Chongkun and Liang, Bin},
  journal={IEEE Robotics and Automation Letters}, 
  title={Single-Actuator Gripper Using an Antagonistic Cable-Driven Differential Mechanism for Adaptive Fixed-Position Grasping}, 
  year={2026},
  volume={11},
  number={6},
  pages={7284-7291},
  doi={10.1109/LRA.2026.3686682}}

@INPROCEEDINGS{Hoang2024,
  author={Hoang, Viet Duong and Falk Nyboe, Frederik and Malle, Nicolaj Haarhøj and Ebeid, Emad},
  booktitle={2024 IEEE International Conference on Robotics and Automation (ICRA)}, 
  title={Autonomous Overhead Powerline Recharging for Uninterrupted Drone Operations}, 
  year={2024},
  volume={},
  number={},
  pages={13551-13557},
  doi={10.1109/ICRA57147.2024.10611522}}

@INPROCEEDINGS{Adikaram2026,
  author={Adikaram, Nimantha and Chandrajith, Lakmina and Abeyratne, Mahen and De Silva, Eranga and Naotunna, Isira and Perera, Asanka},
  booktitle={2026 Moratuwa Engineering Research Conference (MERCon)}, 
  title={A Vision-Guided Hovering System for Autonomous UAV Package Pickup}, 
  year={2026},
  volume={},
  number={},
  pages={1076-1081},
  doi={10.1109/MERCon71835.2026.11691479}}

@ARTICLE{Lee2021,
  author={Lee, Loong Yi and Syadiqeen, Omar Ali and Tan, Chee Pin and Nurzaman, Surya G.},
  journal={IEEE Robotics and Automation Letters}, 
  title={Closed-Structure Compliant Gripper With Morphologically Optimized Multi-Material Fingertips for Aerial Grasping}, 
  year={2021},
  volume={6},
  number={2},
  pages={887-894},
  doi={10.1109/LRA.2021.3052420}}

\end{document}